\documentclass{article}

\usepackage{preprint}
\pdfoutput=1
\usepackage[utf8]{inputenc} 
\usepackage[T1]{fontenc}    
\usepackage{hyperref}       
\usepackage{url}            
\usepackage{booktabs}       
\usepackage{amsfonts}       
\usepackage{nicefrac}       
\usepackage{microtype}      
\usepackage{lipsum}
\usepackage{wrapfig}

\usepackage{booktabs} 
\usepackage{subfig}
\usepackage{amsmath,amssymb}
\usepackage{mathtools}
\usepackage[toc,page]{appendix}
\usepackage{enumitem}
\usepackage{graphics}
\usepackage{graphicx}
\usepackage{algorithm}
\usepackage{algorithmicx}
\usepackage[noend]{algpseudocode}
\algnewcommand{\LineComment}[1]{\State \(//\) #1}

\usepackage[font=small]{caption}

\title{Recognizing Long Grammatical Sequences using Recurrent Networks Augmented with an External Differentiable Stack}

\author{
  Ankur Mali \\
  The Pennsylvania State University \\
  State College, PA 16801 \\
  \texttt{aam35@psu.edu} \\
  \And 
  Alexander Ororbia \\
  Rochester Institute of Technology \\
  Rochester, NY 14623 \\
  \texttt{ago@cs.rit.edu} \\ 
  \AND 
  Daniel Kifer \\
  The Pennsylvania State University \\
  State College, PA 16801 \\
  \texttt{duk17@psu.edu} \\
  \And
  C. Lee Giles \\
  The Pennsylvania State University \\
  State College, PA 16801 \\
  \texttt{clg20@psu.edu} \\
}

\begin{document}
\maketitle

\begin{abstract}
Recurrent neural networks (RNNs) are a widely used deep architecture for sequence modeling, generation, and prediction. Despite success in applications such as machine translation and voice recognition, these stateful models have several critical shortcomings. Specifically, RNNs generalize poorly over very long sequences, which limits their applicability to many important temporal processing and time series forecasting problems. For example, RNNs struggle in recognizing complex context free languages (CFLs), never reaching 100\% accuracy on training. One way to address these shortcomings is to couple an RNN with an external, differentiable memory structure, such as a stack. However, differentiable memories in prior work have neither been extensively studied on CFLs nor tested on sequences longer than those seen in training. The few efforts that have studied them have shown that continuous differentiable memory structures yield poor generalization for complex CFLs, making the RNN less interpretable. In this paper, we improve the memory-augmented RNN with important architectural and state updating mechanisms that ensure that the model learns to properly balance the use of its latent states with external memory. Our improved RNN models exhibit better generalization performance and are able to classify long strings generated by complex hierarchical context free grammars (CFGs). We evaluate our models on CGGs, including the Dyck languages, as well as on the Penn Treebank language modelling task, and achieve stable, robust performance across these benchmarks. Furthermore, we show that only our memory-augmented networks are capable of retaining memory for a longer duration -- up to strings of length $160$. 

\end{abstract}

\section{Introduction}
\label{sec:intro}
Recurrent Neural Networks (RNNs) are powerful, stateful models that capture temporal dependencies in sequential data. They have been widely used in solving complex tasks, such as machine translation \cite{kalch13,bahdanau14,suts14}, language modeling \cite{mikolovrnn10,sunder12,ororbia2017learning}, multimodal language modelling \cite{kiros2014multimodal,ororbia2019like} and speech recognition \cite{graves13}. In theory, first order RNNs with the desired weights (arranged as 2D weight matrices) and infinite precision have been shown to be as powerful as a pushdown automaton (PDA) \cite{gru2019pda,kremer1995computational,frasconi1996computational} and can be considered to be computationally universal models \cite{siegelmann94,siegelmann95}. However, a higher order model (such as second order) offers further representational capability -- \cite{omlin1996constructing,horne1994bounds,carrasco2000stable} proved that any finite state machine can be stably constructed in a $2$nd order, or tensor, RNN.

Languages that are recognizable by first order RNNs, as well as the limits of their computational power, are still being explored. \cite{Merrill19} recently theoretically demonstrated the expressiveness of RNNs with finite precision under asymptotic conditions. They showed that RNNs and those that use gated recurrent units (GRUs) work reasonably well on regular grammars while long short-term memory (LSTM) can recognize real-time counter languages such as $a^nb^n$ or $a^nb^nc^n$ as well as variations. Empirically, LSTM RNNs have demonstrated the capability to recognize simple context free languages (such as $a^nb^n$) and context-sensitive languages (such as $a^nb^nc^n$) by using a complex counting mechanism \cite{lstmcfg,weiss18} where as GRUs struggle to perform dynamic counting \cite{weiss18}. 
While some of the languages from the Chomsky hierarchy \cite{chomsky2002syntactic} can be modelled based on counting, counting does not capture the actual structural property underlying natural languages -- formal languages are hierarchically structured \cite{chomsky2002syntactic,chomsky1962context}. As a result, hierarchy creates a deeply nested structured yielding what is known as the strictly context free languages (CFLs). 
To recognize these CFLs, a simple counting mechanism is not sufficient and, as a result, a stack is necessary in order to correctly recognize these grammars. Based on formal language theory, Dyck languages $D_n$, where $n>1$, are similar to CFLs\cite{nivat1970some}.
\cite{Lstmdynamiccounting} showed that first order RNNs, e.g., GRU, LSTM, perform well on Dyck languages ($D_1)$ and its various combinations, when $n=1$. However, these models struggle whenever $n$ is increased. In theory, a pushdown automaton with a stack can correctly recognize any CFG which is also known to be the homomorphic image of a regular $D_n$ language \cite{chomsky1959algebraic,chomsky1962context}. 

Prior work has focused on using first order RNNs with and without external memory to learn Dyck languages up to length $2$ \cite{Zaremba16,hao2018context,deleu16} but none of them have achieved reasonable performance. However recently \cite{suzgun2019memoryaugmented} showed the effective mechanism for training memory augmented RNNs and thus achieving reasonable performance on Dyck languages. 
In contrast, recent work has focused on a different class of RNNs, often referred to as higher order tensor RNNs, which has demonstrated their ability to learn CFLs, even $D_2$ languages, on long sequences that are extend beyond the training distribution \cite{mali2019neural}. These models consist of high order tensors and, with finite precision, are theoretically capable of recognizing a PDA of any length. These models are furthermore more interpretable and work across a wide variety of grammars \cite{mali2019neural,das1993using,omlin1996extraction}. However, due to their computational complexity and also difficulty in training, higher order RNNs are often not used in practice.
In this paper, we focus on investigating the limitations of first order RNNs with finite precision, with and without external memory, for the task of recognizing CFLs of lengths $n>1$. Furthermore, we propose several new, improved memory-augmented RNN models, a class we label as the DiffStk-RNN family of RNN models, that can effectively recognize long CFLs.
The contributions of this paper are the following:
\begin{itemize}
\item Five new, improved stack-augmented RNNs are introduced and within a non-traditional framework of next-step prediction, a schema that we show facilitates stronger, iterative error-correction for improved training of CFG recognition models. 
We show that approximating higher order RNNs with stack memory offers the best results when classifying long CFL strings with reduced computation time.
\item The performance of first order RNNs on complex, long string CFLS are investigated for the first time. We show the importance of using negative string examples when training for CFG recognition.
Our experiments show that our models, with only finite precision, can emulate the dynamics of a PDA to effectively learn long string CFLs, mainly $D_>1$ languages, as well as perform well on the Penn Treebank langauge modeling dataset.
\item An efficient scheme is proposed for training stack augmented RNNs, which helps to preserve memory over longer string lengths. In addition, we demonstrate the importance of noise and the need for better handling of NO-OP operations on an RNN's stack when attempting to preserve memory over longer time spans.
\end{itemize}

\section{Related Work}
\label{sec:related_work}
Historically, grammatical inference \cite{gold1967language} has been at the core of formal language theory and could be considered to be fundamental in understanding important properties of natural languages. 
A summary of much of the theoretical work done in formal languages can be found in \cite{de2010grammatical}.
Applications of formal language work have led to the development of methods for predicting and understanding sequences in diverse areas, such as financial time series, genetics and bioinformatics, and software data exchange \cite{giles2001noisy,wieczorek2016use,exler2018grammatical}. 
Many artificial neural network (ANN) models take the form of a first order recurrent neural network (RNN) and are taught to learn context free and context-sensitive counter languages \cite{lstmcfg,boden2000context,tabor2000fractal,wiles1995learning,sennhauser2018evaluating,nam2019number,wang2019stateregularized,arabshahi2019memory,cleeremans1989finite,kolen1994recurrent,cleeremans1989finite,weiss2017extracting}. However, from a theoretical perspective, RNNs augmented with an external memory have historically been shown to be more capable of recognizing context free languages (CFLs), such as with a discrete stack \cite{das1993using,pollack1990recursive,sun1997neural}, or, more recently, with various differentiable memory structures \cite{joulin2015inferring,grefenstette2015learning,graves2014neural,kurach2015neural,zeng1994discrete,hao2018context,yogatama2018memory,graves2016hybrid,le2019neural,mali2019neural}. Despite the positive recognition results, prior work on CFLs was unable to show model generalization beyond the training dataset, highlighting a troubling difficulty in training these kinds of networks. In contrast, \cite{zeng1994discrete} and \cite{lstmcfg} demonstrated that their RNN models could potentially generalize beyond the training dataset on simple CFLs. 

Recent work on differentiable stacks (the stackRNN) \cite{joulin2015inferring}, which is closely related to this work, tested models on real natural language modeling tasks and on learning simple algorithmic patterns. These models were able to solve problems which required counting as well as memorization. Other work related to differentiable memory \cite{grefenstette2015learning} was motivated by the neural pushdown automaton, i.e., the NNPDA \cite{das1992learning,das1993using, mozer1993connectionist,sun1997neural}, which extended RNNs to use an unbounded external differentiable memories such as stacks, queues, and doubly-linked lists. However, the computational limitations of these models were, unfortunately, only explored on synthetic datasets. Furthermore, none of the these models were tested on complex CFLs and the internal hidden representations were never analyzed in an effort to understand the internal network dynamics. Other memory-augmented models have also been proposed to solve CFLS, such as neural random access memory (NRAM) \cite{kurach2015neural} and the neural Turing machine (NTM) \cite{graves2014neural}, but these models face a plethora of instability issues due to their structure and are quite difficult and expensive to train for real world problems. While NTMs and LSTMs were tested on Dyck ($D_1$) languages over longer sequences and shown to be able to perfectly learn $D_1$, these results, however, are limited given that a $D_1$ language can also be easily learned by a counting model or counting automaton. 
To truly test the generalization ability of first order RNNs and other memory-augmented models, one should test on more complex CFLs such as the Dyck languages. 
Recently, a much more powerful and interpretable model, known as the neural state pushdown neural network (NSPDA), was proposed, demonstrating strong generalization on the Dyck ($D_2$) languages \cite{mali2019neural} while first order RNNs struggled when attempting to recognize more complex CFLs.
Unfortunately, due to the fact that the NSPDA makes heavy use of higher order weights, it would still be an expensive model to train on real world datasets, much like its ancestor, the NNPDA.  To complement the rich body of historical work that has aimed to explore RNN performance on CFL recognition \cite{mali2019neural,giles1992learning,zeng1994discrete,frasconi1996computational,joulin2015inferring}, we offer additional evidence of the limitations of first order RNNs in recognizing Dyck languages ($D_>1)$, strongly demonstrating the need for memory structures when attempting to learn recursive languages. We circumvent the limitations of the NSPDA's expensive training by crafting five different models and their training schemes for recognizing long strings generated by complex CFLs.

\section{Models Augmented with an External Stack}
\label{sec:stack_models}
In this section, we describe various stack-augmented RNNs. Our models are motivated by neural systems designed to emulate state machines with pushdown dynamics \cite{das1993using,mali2019neural,joulin2015inferring,grefenstette2015learning,hao2018context,arabshahi2019memory,suzgun2019memoryaugmented}. Unlike these previous models, ours are much simpler and can be trained and tested on real world tasks.

In this study, we approach the problem of CFL recognition a bit differently compared to how it has been approached historically. Classically, an RNN designed for recognizing CFLs first encodes the entire string sequence and then subsequently classifies it as correct or incorrect, i.e., the input to the recognition classifier was the RNN's final, computed hidden vector summary of the sequence. 
In contrast, we adopt a hybrid, next-step prediction scheme which entails: 1) predicting the recognition label $y$ of the string at each time step, measuring efficacy with a mean squared error (MSE) loss 
, and, 2) predicting the symbol $\mathbf{x}_t$ given a history of symbols observed thus far $\mathbf{x}_{<t}$ (using a cross-entropy loss as is characteristic of RNN language models). Note that $\mathbf{x}_t \in \{0,1\}^{Vx1}$ is a one-hot encoding of the symbol at step $t$ in a sequence of length $T$ and $V$ is the total number of unique symbols in the dataset/corpus. Assuming the CFL model's prediction (at step $t$) of the recognition label is $\hat{y}_t \in [0,1]$ (a scalar) and its prediction of the next token is $\mathbf{\hat{x}}_t$, then the complete loss for a CFL sample sequence string becomes:
\begin{align*}
    \mathcal{L}(\Theta) = \sum^T_{t=1} \bigg[ \sum_i -(\mathbf{x}_t \otimes \log(\mathbf{\hat{x}}_t))[i] \bigg] + \frac{1}{2}(\hat{y}_t - y)^2
\end{align*}
where $i$ indicates retrieval of the $i$th scalar value in a vector.
Our hybrid, next-step prediction scheme was motivated by the intuition that next step prediction might facilitate a better training scheme for CFL recognition and allow the RNN to recover from previous mistakes made when iteratively recognizing the sequence. In other words, next-step prediction provides a stronger form of error correction, as was also discovered when training first order recurrent nets \cite{lstmcfg} and also to train complex, higher-order neural models \cite{mali2019neural}. From a model training point-of-view, having a next-step prediction scheme also provides a wealth of gradient signals, since predictions are being made and corrected at each time step, that help combat the vanishing gradient problem \cite{glorot2010understanding} that plagues training RNN architectures.
This is particularly important for the CLF recognition task given that models are expected to operate and learn from very long pattern sequences.

Ideally, since we want all of our networks to operate like PDAs, we formulate in the spirit of classical pushdown networks. As a result, we first start by defining an $M$-state PDA as a 7-tuple $(Q,\Sigma,\Gamma, \delta,{q^0},\bot,F)$ where:
\begin{itemize}
    \item $\Sigma = \{a^1,\cdots,a^l,\cdots,a^L\}$ is the input alphabet
    \item $Q = \{s^1,\cdots,s^m,\cdots,s^M\}$ is the finite set of states
    \item  $\Gamma$ is known as stack alphabet (a finite set of tokens)
    \item ${q^0}$ is the start state
    \item  $\bot$ is the initial stack symbol
    \item $F \subseteq Q$ is the set of accepting states
    \item $\delta \subseteq Q \times (\Sigma \cup )| \times \Gamma \rightarrow Q \times \Gamma^*)$ is the state transition.
\end{itemize}


\subsection{The DiffStk-RNN Pushdown Network}
\label{sec:stackrnn_pdn}
Our pushdown network, which we will refer to as ``Differentiable Stack-RNN'', or DiffStk-RNN (to distinguish it from the original model ``StackRNN''\cite{joulin2015inferring}) and \cite{suzgun2019memoryaugmented}, is concretely instantiated as an RNN that learns to control an external stack data structure. The RNN itself consists of an input layer, a hidden layer with recurrent connections, and an output layer. The RNN process inputs in a sequence step by step, computing a hidden state:
\begin{align}
    \mathbf{z}_t = f_1(U \mathbf{x}_t + R \mathbf{z}_{t-1}) \label{eqn:RNN_update}
\end{align}
where $f_1(\circ)$ is hyperbolic tangent (tanh) activation function, specifically, the scaled variant of it \cite{lecun2012efficient},
\begin{align}
    f_1(\circ) = 1.7519 \times tanh(2/3 \times \circ)
\end{align}{}
applied coordinate-wise. Note that U is a $m \times d$ token embedding matrix and R is a $m \times m$ recurrent weight matrix.  The number of different tokens is $d$ while $m$ is the number of hidden units in the state $\mathbf{z}_t$. Based on its current state at step $t$, the RNN outputs a probability vector $\mathbf{y}_t$ for the next token as follows:
\begin{align}
\mathbf{y}_t = softmax(V \mathbf{z}_t) = \frac{\exp(V\mathbf{z}_t)}{\sum_i \exp(V\mathbf{z}_t)[i]}
\end{align}
where $V$ is $d \times m$ output matrix. Despite its success in natural language processing, this architecture, as well as its variants, are unable to recognize complex CFLs. If we were to, however, equip the network with an external memory, then it would be theoretically possible to generally recognize a CFL \cite{giles1992learning,mali2019neural,zeng1994discrete}. Hence, we augment the RNN with a stack, which serves as an external persistent memory structure which has only its topmost element readily accessible to the RNN controller. In this case, such an RNN has basic three operations: 
1) PUSH: adds elements, 2) POP: removes elements, and 3) NO-OP (NoOP): does nothing. 
We want the stack to carry information to the hidden layer of an RNN (acting as the controller). While the original StackRNN  does this according to the following equation (biases omitted for clarity):
\begin{align}
\mathbf{z}_t = f_1(U \mathbf{x}_t + R\mathbf{z}_{t-1})
\end{align}{}
our DiffStk-RNN does with an additional term added to the hidden state update equation, plus others tricks such as better handling of no-op and negative sampling to make it stable. 
Specifically, we compute the next hidden state of our model according to the following:
\begin{align}
\mathbf{\hat{z}}_{t-1} = \mathbf{z}_{t-1} + P \mathbf{S}_{t-1}^0,\quad \mathbf{z}_t = f_1(U \mathbf{x}_t + R\mathbf{\hat{z}}_{t-1})
\end{align}
where P is $ m \times 3$ 
recurrent matrix and $\mathbf{S}_{t-1}[0]$ is the top-most element of stack $\mathbf{S}$. The formula above demonstrates how the stack-augmented term is integrated into the hidden state calculation.
To specify the stack itself, we start by denoting $a_t$ as a 3-dimensional variable representing each action taken on the stack, which is dependent on hidden state $\mathbf{z}_t$
\begin{align}
\mathbf{a}_t = softmax(A \mathbf{z}_t) 
\end{align}
where $A$ is a $3 \times m $ matrix. We store the top element of the stack at position $0$, with value $\mathbf{S}_t[0]$, via the following:
\begin{align}
    \mathbf{S}_t[0] = \mathbf{a}_t[PUSH]\sigma (D\mathbf{z}_t) + \mathbf{a}_t[POP]\mathbf{S}_{t-1}[1] + \mathbf{a}_t[NoOP]\mathbf{S}_{t-1}[0]
\end{align}
where the symbols PUSH, POP, and NoOP correspond to the unique integer indices $0$, $1$, and $2$ that access the specific action value in their respective slot.
$D$ is a $1 \times m$ matrix and $\sigma(x) = 1/(1+exp(-x))$ is the logistic sigmoid function. If $\mathbf{a}_t[PUSH] = 1$ we add element to the top of the stack and if $\mathbf{a}_t[POP] = 1$ we remove the element at top of the stack and move the stack upwards.
Similarly for elements stored at depth $i>0$ in the stack, the rule is as follows:
\begin{align}
    \mathbf{S}_t[i] = \mathbf{a}_t[PUSH]\mathbf{S}_{t-1}[i-1] + \mathbf{a}_t[POP]\mathbf{S}_{t-1}[i+1] \mbox{.}
\end{align}
Note that for more complex hidden state functions used to compute $\mathbf{z}$ (such as an LSTM state function), integration with a stack is the same as described in this section.

\textbf{Adding Noise to the Hidden Weights:} Noise plays a crucial role in regularizing RNNs. Besides regularization, incorporating it has also been shown to improve the model's forecasting horizon yielding better generalization \cite{mali2019neural,mali2019sibling,noh2017regularizing,jim1996analysis,goodfellow2016deep}. Motivated by this result, we modify the state update equation further as follows:
\begin{align}
    \mathbf{\hat{z}}_{t-1} = \mathbf{z}_{t-1} + P \mathbf{s}_{t-1}^0 + \mathbf{\epsilon},\quad  \forall i, \mathbf{\epsilon}[i] \sim \mathcal{N}(\mu, \sigma^2)
\end{align}
where  $\mu$ (mean) and $\sigma^2$ (variance) are user-set meta-parameters to control the strength of the additive Gaussian regularization noise. Based on our experiments, we found that injecting a small amount of this type of noise into the state improve the RNN recognition model's generalization even though it does slow down model convergence. Table \ref{noiseperformance} supports over claim. \footnote{Other regularization schemes were tried, such as drop-out, batch normalization, and layer normalization. However, these resulted in poorer generalization performance and thus we focus in this study on state-injected additive noise.}

\begin{table}[!t]
    \centering
    \begin{tabular}{l|r|r||r|r}
    \multicolumn{1}{l}{}&\multicolumn{2}{|c}{\begin{tabular}[x]{@{}c@{}}\textbf{$With Noise$}\\\end{tabular}}&\multicolumn{2}{|c}{\begin{tabular}[x]{@{}c@{}}\textbf{$W/o Noise$}\\\end{tabular}}\tabularnewline
    \multicolumn{1}{l}{\textbf{RNN Models}}&\multicolumn{1}{|c}{\begin{tabular}[x]{@{}c@{}}\textbf{Mean}\\\end{tabular}}&\multicolumn{1}{|c}{\begin{tabular}[x]{@{}c@{}}\textbf{Best}\\\end{tabular}}&\multicolumn{1}{|c}{\begin{tabular}[x]{@{}c@{}}\textbf{Mean}\\\end{tabular}}&\multicolumn{1}{|c}{\begin{tabular}[x]{@{}c@{}}\textbf{Best}\\\end{tabular}}\tabularnewline
\hline
        \textit{RNN} & $0.2$ & $0.4$ & $0.0$ & $0.0$  \tabularnewline
        \textit{LSTM} & $1.40$ & $4.00$ & $1.00$ & $3.85$  \tabularnewline
        \textit{StackRNN} & $70$ & $100$ & $70$ & $99.95$  \tabularnewline
        \textit{DiffStk-RNN (Ours)} & $99.99$ & $100$ & $97.58$ & $99.00$  \tabularnewline
        \textit{DiffStk-MRNN (Ours)} & $\textbf{99.00}$ & $\textbf{100}$ & $97.20$ & $99.99$  \tabularnewline
        \textit{DiffStk-MIRNN (Ours)} & $97.25$ & $99$ & $97.00$ & $99$  \tabularnewline
    \end{tabular}
    \caption{ Percentage of correctly classified strings of various RNNs, trained with and without noise, on $D_2$ language (averaged over $10$ trials), with mean and best accuracy measurements reported for each model. The test set used for all experiments had short samples mixed with long ones up to length $T = 102$, containing both positive and negative sample strings.}
    \label{noiseperformance}
\end{table}

\textbf{Carry Forward State Update:} Whenever there exists a symbol that leads to more than one No-OP operation on the stack, we propose carrying forward the previous hidden state. Empirically, we found that this helps whenever the RNN encounters too many No-OPs in the input strings. This can be considered as approximation of a ``reject'' state, which have been shown to stabilize the learning process of higher order RNNs \cite{das1993using,nndpa1998sun}. We keep a counter which indicates how many No-OPs the network has seen so far, which is used to trigger the update equation. 
With this counter-based trigger integrated, the full hidden state update equation can then be written as follows:
\begin{align}
  \mathbf{z}_t = u_t f_1(U \mathbf{x}_t + R \mathbf{\hat{z}}_{t-1}) + (1-u_t) \mathbf{\hat{z}}_{t-1}, \quad
  u_t = \bigg \{ 
    \begin{array}{cc}
      u_t = 0, and ct>1 &  \mbox{if NoOP} \\
      u_t = 1, & \mbox{otherwise} \\
    \end{array}
\end{align}
where $u_t$ is a scalar: $0$ indicates that the RNN is to copy the previous state while $1$ indicates that the state is to be updated/overwritten and ct is the counter which keeps track of No-oP operations \footnote{We maintain counter variable just to keep track of No-op and is just used to facilitate decision for RNN and is based on $S_t$ }.
In our analysis, presented in Table \ref{stateperformance}, we show that models using noise in conjunction with carry forward states work better than models that do not. We report accuracy over $10$ trials on a $D_2$ grammar. Observe that our approach consistently demonstrates improved memory retention over longer pattern sequence lengths. 

\begin{table}[!t]
    \centering
    \begin{tabular}{l|r|r||r|r}
    \multicolumn{1}{l}{}&\multicolumn{2}{|c}{\begin{tabular}[x]{@{}c@{}}\textbf{$W changes$}\\\end{tabular}}&\multicolumn{2}{|c}{\begin{tabular}[x]{@{}c@{}}\textbf{$W/o changes$}\\\end{tabular}}\tabularnewline
    \multicolumn{1}{l}{\textbf{RNN Models}}&\multicolumn{1}{|c}{\begin{tabular}[x]{@{}c@{}}\textbf{Mean}\\\end{tabular}}&\multicolumn{1}{|c}{\begin{tabular}[x]{@{}c@{}}\textbf{Best}\\\end{tabular}}&\multicolumn{1}{|c}{\begin{tabular}[x]{@{}c@{}}\textbf{Mean}\\\end{tabular}}&\multicolumn{1}{|c}{\begin{tabular}[x]{@{}c@{}}\textbf{Best}\\\end{tabular}}\tabularnewline
\hline
        \textit{RNN} & $0.2$ & $0.4$ & $0.0$ & $0.0$  \tabularnewline
        \textit{LSTM} & $1.40$ & $4.00$ & $1.00$ & $3.85$  \tabularnewline
        \textit{StackRNN} & $70$ & $100$ & $70$ & $99.95$  \tabularnewline
        \textit{DiffStk-RNN (Ours)} & $99.99$ & $100$ & $97.58$ & $99.00$  \tabularnewline
        \textit{DiffStk-MRNN (Ours)} & $\textbf{100}$ & $\textbf{100}$ & $97.20$ & $99.99$  \tabularnewline
        \textit{DiffStk-MIRNN (Ours)} & $98.20$ & $100$ & $97.00$ & $99$  \tabularnewline
    \end{tabular}
    \caption{Percentage of correctly classified strings of various RNNs with and without noise as well as carry forwarding state on $D_2$ language (averaged over $10$ trials). We report the mean and best accuracy for each of the models. The test set had short samples mixed with long ones up to length $T = 102$, with both positive and negative samples, for all experiments}
    \label{stateperformance}
\end{table}

\textbf{Negative Sampling:} Based on extensive preliminary experimentation, we found that it is crucially important to provide enough negative samples to a CFG-recognition network during both the training and testing phases (whether using our next-step prediction scheme or the classical scheme or predicting the label at the end). While the rough ratio of positive to negative samples can be controlled when first sampling a target CFG to create a training dataset, most of the negative samples produced by this process are simply too different from the positive cases, or rather, are simply too easy to distinguish from positive cases. In order to truly improve the generalization ability of our RNN models, we seek to mix with sampled negative cases some number of ``difficult'' samples, which are similar/close to positive strings but would still be rejected by the target CFG.

In order to facilitate the generation of these kinds of negative samples, we utilize an oracle which recognizes any given language. We then randomly swap around $1$ or $3$ characters in the input strings and use the oracle to check whether or not the generated grammar is negative (generating the correct rejection label). 
We then generate and mix in these synthetic negative samples to the dataset while still preserving a balance between the number of positive and negative examples in the original CFG recognition training set. Specifically, we generate a new number of difficult negative samples (Ns) and swap out the same number from the original pool of negative strings. Specifically, prior to training, we stochastically set the number to be somewhere between $15$-$30$\% of the total number of original negative samples, generate these new examples based on the scheme above, and swap out the same number of randomly chosen original negative strings with the new, more difficult ones.

Based on our experiments, we find that training with our form of negative sampling does indeed improve the performance of an RNN model over longer strings.    


\subsection{ DiffStk-LSTM}
\label{sec:stack_lstm}
The  DiffStk-LSTM is similar to the  DiffStk-RNN but with the exception that it contains additional gating components that help it to better preserve memory over long time spans. There are several variants of LSTM \cite{lstmcfg} and we will present the variant we used in our experiments for this study.
The hidden state for our LSTM is computed as follows:
\begin{align}
    \mathbf{\hat{z}}_t = U \mathbf{x}_t + R \mathbf{z}_{t-1}
\end{align}
where U is a $x \times z$ matrix and R is a $z \times z$ matrix. An LSTM also consists of $3$ gating units -- an input gate $\mathbf{i}$, an output gate $\mathbf{o}$, and a forget gate $\mathbf{f}$, which all sport recurrent and feedforward connections:
\begin{align}
    \mathbf{i}_t &= \sigma(U_i \mathbf{x}_t + R_i \mathbf{z}_{t-1})\\
    \mathbf{o}_t &= \sigma(U_o \mathbf{x}_t + R_o \mathbf{z}_{t-1})\\
    \mathbf{f}_t &= \sigma(U_{f} \mathbf{x}_t + R_{f} \mathbf{z}_{t-1})
\end{align}
where $\sigma$ is the logistic sigmoid. The internal state and output of the hidden state are calculated as follows:
\begin{align}
    \mathbf{c}_t &= \mathbf{f}_t \odot \mathbf{c}_{t-1} + \mathbf{i}_t \odot f_1(\hat{\mathbf{z}_t}) \\
    \mathbf{z}_t &= f_1(\mathbf{c}_t) \odot \mathbf{o}_t
\end{align}
The ability of LSTM to control how information is stored in each of its cells has proven to be useful in many applications. 

\subsection{DiffStk-MRNN}
\label{sec:stack_mrnn}
The multiplicative RNN (MRNN) \cite{krause2016multiplicative} is similar in spirit to second order tensor RNNs \cite{giles1990higher,watrous1992induction,forcada1995learning,ziemke1996towards,kremer1998constrained,carrasco1995second}. It uses a factorized hidden-to-hidden transition matrix in place of the normal RNN hidden-to-hidden matrix ($R$ or $W_{zz}$). The MRNN can be formulated to compute an intermediate state $\mathbf{m}_t$ as follows:
\begin{align}
    \mathbf{m}_t = (W_{mx} \mathbf{x}_t) \cdot (W_{mz} \mathbf{z}_{t-1})
\end{align}
\begin{align}
    \mathbf{\hat{z}}_t = W_{zm} \mathbf{m}_t + W_{zx} \mathbf{x}_t
\end{align}{}
with the final hidden state update calculated as follows:
\begin{align}
    \mathbf{\hat{z}}_t = f_1(U \mathbf{x}_t + \hat{R} \mathbf{\hat{z}}_{t-1})
\end{align}
where $\hat{R}$ is a $z \times m$ matrix. 
This can be shown to approximate an NSPDA with second order weights \cite{mali2019neural,das1993using,nndpa1998sun}. 

\subsection{DiffStk-MLSTM}
\label{sec:stack_lstm}
Motivated by the success of tensor RNNs on complex CFLs, we integrated a multiplicative LSTM (MLSTM), a hybrid model which combines a factorized hidden-to-hidden transition (which is an approximation of a second order RNN) with our differentiable stack. An MRNN's intermediate state ($\mathbf{m}_t$) can easily be connected to the LSTM gating units. This gives the model not only more expressive cells but also allows it to operate similarly to a pushdown automaton. The resulting formulation of the model is as follows:
\begin{align}
    \mathbf{m}_t = (U_m \mathbf{x}_t) \odot (R_m \mathbf{z}_{t-1})\\
    \mathbf{\hat{z}}_t = U \mathbf{x}_t + R_z \mathbf{m}_t) \\
    \mathbf{i}_t = \sigma(U_i \mathbf{x}_t + R_{im} \mathbf{m}_t)\\
    \mathbf{o}_t = \sigma(U_o \mathbf{x}_t + R_{om} \mathbf{m}_t)\\
    \mathbf{f}_t = \sigma(U_{f} \mathbf{x}_t + R_{fom} \mathbf{m}_t)
\end{align}
where $U_m$ is $m \times x$ matrix, $R_m$ is a $m \times z$ matrix, $R_{zm}$ is a $z \times m$ matrix, $R_{im}$ is a $i \times m$ matrix, $R_{om}$ is a $o \times m$ matrix and $R_{fom}$ is a $f \times m$ matrix. In all of our experiments, we set the dimension of the state $\mathbf{z}$ and $\mathbf{m}$ to be similar. Integration of a stack with the MLSTM cell is similar to the formulation provided earlier in  DiffStk-RNN section. 

\subsection{DiffStk-MIRNN}
\label{sec:stack_mirnn}
Multiplicative integration RNN (MIRNN) \cite{wu2016multiplicative} is yet another approximation of higher order/tensor RNNs \cite{giles1990higher,watrous1992induction,forcada1995learning,ziemke1996towards,kremer1998constrained,carrasco1995second}. This model uses a Hadamard product $\odot$ instead of a sum operation $+$ in the standard RNN update rule. Hence, this new formulation allows for a potentially more powerful for the RNN to update its hidden state without introducing any extra parameters. This change can be written in the following manner:
\begin{align}
    \mathbf{z}_t = f1(U \mathbf{x}_t \odot R \mathbf{z}_{t-1}) \mbox{.}
\end{align}
These models have the capacity to retain memory over longer time spans and their approximate second order connections add extra expressiveness that prove useful when learning to recognize complex CFLs. Integration of a stack with the MI-RNN cell is also similar to the previous models discussed.

\section{Experiments}
\label{sec:experiments} 
To evaluate our models and investigate the limitations of first order RNNs, we tested all models on recursive languages, including the Dyck languages, which require a stack-like structure as opposed to a simple counting mechanism in order to ensure successful recognition. 
In addition, we also tested our models on a real-world problem, i.e., the Penn Treebank language modeling benchmark. 
To our knowledge this is the first work to extensively compare a large variety of first order RNNs on complex CFLs task. We followed two training processes, one was sequential \cite{lstmcfg,mikolovrnn10}, which is widely used in natural language processing related tasks, while the other was based on incremental learning \cite{giles1992learning,mali2019neural} or curriculum learning \cite{bengio2009curriculum}, which is a popular training scheme for tensor RNNs. We compare these two variants in Table \ref{incrementperformance} and show that sequential prediction has a slight advantage over incremental learning.

\begin{table}[!t]
    \centering
    \begin{tabular}{l|r|r||r|r}
    \multicolumn{1}{l}{}&\multicolumn{2}{|c}{\begin{tabular}[x]{@{}c@{}}\textbf{$Sequential$}\\\end{tabular}}&\multicolumn{2}{|c}{\begin{tabular}[x]{@{}c@{}}\textbf{$Incremental$}\\\end{tabular}}\tabularnewline
    \multicolumn{1}{l}{\textbf{RNN Models}}&\multicolumn{1}{|c}{\begin{tabular}[x]{@{}c@{}}\textbf{Mean}\\\end{tabular}}&\multicolumn{1}{|c}{\begin{tabular}[x]{@{}c@{}}\textbf{Best}\\\end{tabular}}&\multicolumn{1}{|c}{\begin{tabular}[x]{@{}c@{}}\textbf{Mean}\\\end{tabular}}&\multicolumn{1}{|c}{\begin{tabular}[x]{@{}c@{}}\textbf{Best}\\\end{tabular}}\tabularnewline
\hline
        \textit{RNN} & $0.2$ & $0.4$ & $0.0$ & $0.0$  \tabularnewline
        \textit{LSTM} & $1.40$ & $4.00$ & $1.00$ & $3.85$  \tabularnewline
        \textit{StackRNN} & $70$ & $100$ & $70$ & $99.95$  \tabularnewline
        \textit{DiffStk-RNN (Ours)} & $99.99$ & $100$ & $97.58$ & $99.00$  \tabularnewline
        \textit{DiffStk-MRNN (Ours)} & $\textbf{100}$ & $\textbf{100}$ & $97.20$ & $99.99$  \tabularnewline
        \textit{DiffStk-MIRNN (Ours)} & $98.20$ & $100$ & $97.00$ & $99$  \tabularnewline
    \end{tabular}
    \caption{Percentage of correctly classified strings of various RNNs trained under incremental vs sequential schemes on $D_2$ language (measurements averaged over 10 trials). We report the mean and best accuracy for each model. The test set used for all experiments had short samples mixed with long ones up to length $T = 102$ containing both positive and negative sample strings.}
    \label{incrementperformance}
\end{table}

\subsection{Training}
\label{sec:training} 
All RNNs trained were design to have a single layer with $8$ hidden units. RNN weights were adjusted using gradients computed via back-propagation through time (BPTT) (with a look-back that extended $50$ steps). Gradients were hard-clipped to have a maximum magnitude of $15$ in order to ensure that the gradients did not explode explode. Adam \cite{kingma2014adam} was used to update the weights given the gradients using an initial learning rate $2e-3$. A patience schedule was used to adjust the learning rate -- if a gain was not observed in validation within the last three times it was checked, the learning rate was halved. 
We ran our experimental simulations $10$ times, i.e., each run used a unique seed, and we report the mean and best performance accuracy for each model. We optimized all networks over the course of $30$ epochs and report the final training and testing performance for each model.

\section{Datasets}
\label{sec:data}
As noted earlier, $D>_{1}$ denotes all complex context free languages (CFLs) that require a stack-like structure in order for complete and accurate recognition to be possible. Note that a simple counting mechanism (which is what most first-order RNNs implement) is not sufficient to recognize these languages. In light of this, we created various test sets in order to understand the true limits of first order RNNs when recognizing complex CFLs and to test our proposed new models in order to observe potential gains in generalization performance.

\subsection{Context-Free Languages}
\label{sec:cfgs}
We experimented with various complex grammars such as the Dyck languages.
The Dyck languages can be defined in the following manner.
Let $\Sigma = {[,]}$ be the alphabet consisting of symbol [ and ] and let $\Sigma^*$ denote its Kleene closure. A Dyck language is defined as:
\begin{align*}
    \{n \in \Sigma^*  | \mbox{ all prefixes of } n \mbox{ contain no more symbol ']' than symbol '[' \& } cnt(\mbox{'['}, n) = cnt(\mbox{']'}, n)  \}
\end{align*}
where $cnt(\mbox{<symbol>},n)$ is the frequency of <symbol> in $n$.
We can also define Dyck languages via CFGs with a single non-terminal S. From this perspective, the production rule is defined in the following manner.
S $\xrightarrow{}$ $\epsilon | "[" S"]"S$, where $S$ is either empty set or an element of the Dyck languages. A probabilistic context-free grammar for $D_2$ can be written in the following form:
\begin{align}
S \rightarrow \begin{cases} 
(\, S\, ) & \text{with probability } \frac{p}{2} \\
[\, S\, ] & \text{with probability } \frac{p}{2} \\
S\,S & \text{with probability } p_1 \\ 
\epsilon & \text{with probability } 1 - (p+p_1)
\end{cases}
\end{align}
where $\epsilon$, $p$, and $p_1$ are scalar values set externally.

We tested various Dyck languages such as $D_2$, $D_3$, and $D_6$. 
For the $D_2$ and $D_3$ grammars, we created a dataset containing $6230$ training samples (of length $T$ less than or equal to $55$), $1000$ for validation ($20 < T \leq 70$, and $3000$ samples for testing ($55 <  T \leq 102)$. However, for $D_6$, which is a much more complex grammar, we created a dataset containing far more training samples - $15000$ for training with $2000$ set aside for validation and $4000$ for test. The proportions of string lengths in all splits for $D_6$ was kept identical.

Following prior work, we also tested our models on the palindrome language \cite{zeng1994discrete,giles1992learning,lstmcfg,mali2019neural}. 

\begin{table}[!t]
    \centering
    \begin{tabular}{l|r|r||r|r}
    \multicolumn{1}{l}{}&\multicolumn{2}{|c}{\begin{tabular}[x]{@{}c@{}}\textbf{$Train$}\\\end{tabular}}&\multicolumn{2}{|c}{\begin{tabular}[x]{@{}c@{}}\textbf{$Test$}\\\end{tabular}}\tabularnewline
    \multicolumn{1}{l}{\textbf{RNN Models}}&\multicolumn{1}{|c}{\begin{tabular}[x]{@{}c@{}}\textbf{Mean}\\\end{tabular}}&\multicolumn{1}{|c}{\begin{tabular}[x]{@{}c@{}}\textbf{Best}\\\end{tabular}}&\multicolumn{1}{|c}{\begin{tabular}[x]{@{}c@{}}\textbf{Mean}\\\end{tabular}}&\multicolumn{1}{|c}{\begin{tabular}[x]{@{}c@{}}\textbf{Best}\\\end{tabular}}\tabularnewline
\hline
        \textit{RNN} & $9.12$ & $14.02$ & $0.2$ & $0.4$  \tabularnewline
        \textit{LSTM} & $54.00$ & $62.80$ & $1.40$ & $4.00$  \tabularnewline
        \textit{GRU} & $48.00$ & $50.00$ & $1.00$ & $1.02$  \tabularnewline
        \textit{$\Delta$-RNN} & $52.00$ & $60.02$ & $1.50$ & $4.00$  \tabularnewline
        \textit{Stack RNN} & $71$ & $100$ & $70$ & $100$  \tabularnewline
        \textit{DiffStk-RNN\textbf{(Ours)}} & $100$ & $100$ & $99.99$ & $100$  \tabularnewline
        \textit{DiffStk-LSTM\textbf{(Ours)}} & $94.20$ & $100$ & $90$ & $100$  \tabularnewline
        \textit{DiffStk-MRNN\textbf{(Ours)}} & $\textbf{100}$ & $\textbf{100}$ & $\textbf{100}$ & $\textbf{100}$  \tabularnewline
        \textit{DiffStk-MLSTM\textbf{(Ours)}} & $96$ & $100$ & $95.69$ & $100$  \tabularnewline
        \textit{DiffStk-MIRNN\textbf{(Ours)}} & $98$ & $100$ & $98.20$ & $100$  \tabularnewline
    \end{tabular}
    \caption{Percentage of correctly classified strings of various RNNs trained on the $D_2$ language (averaged over $10$ trials). We report the mean and best accuracy for each model. Test set used in all experiments had a mix of short samples and long ones of length up to$T = 102$ containing both positive and negative samples.}
    \label{D2performance}
\end{table}

\begin{table}[!t]
    \centering
    \begin{tabular}{l|r|r||r|r}
    \multicolumn{1}{l}{}&\multicolumn{2}{|c}{\begin{tabular}[x]{@{}c@{}}\textbf{$Train$}\\\end{tabular}}&\multicolumn{2}{|c}{\begin{tabular}[x]{@{}c@{}}\textbf{$Test$}\\\end{tabular}}\tabularnewline
    \multicolumn{1}{l}{\textbf{RNN Models}}&\multicolumn{1}{|c}{\begin{tabular}[x]{@{}c@{}}\textbf{Mean}\\\end{tabular}}&\multicolumn{1}{|c}{\begin{tabular}[x]{@{}c@{}}\textbf{Best}\\\end{tabular}}&\multicolumn{1}{|c}{\begin{tabular}[x]{@{}c@{}}\textbf{Mean}\\\end{tabular}}&\multicolumn{1}{|c}{\begin{tabular}[x]{@{}c@{}}\textbf{Best}\\\end{tabular}}\tabularnewline
\hline
        \textit{RNN} & $9.60$ & $15.02$ & $0.0$ & $0.0$  \tabularnewline
        \textit{LSTM} & $30.88$ & $35.00$ & $0.02$ & $0.04$  \tabularnewline
        \textit{GRU} & $10.00$ & $15.00$ & $0.0$ & $0.0$  \tabularnewline
        \textit{$\Delta$-RNN} & $20.00$ & $21.00$ & $0.02$ & $0.04$  \tabularnewline
        \textit{Stack RNN} & $80$ & $100$ & $69$ & $100$  \tabularnewline
        \textit{DiffStk-RNN\textbf{(Ours)}} & $82.60$ & $100$ & $82.00$ & $100$  \tabularnewline
        \textit{DiffStk-LSTM\textbf{(Ours)}} & $76.75$ & $100$ & $57.50$ & $99.50$  \tabularnewline
        \textit{DiffStk-MRNN\textbf{(Ours)}} & $\textbf{84.02}$ & $\textbf{100}$ & $\textbf{84.00}$ & $\textbf{100}$  \tabularnewline
        \textit{DiffStk-MLSTM\textbf{(Ours)}} & $77$ & $100$ & $59.00$ & $100$  \tabularnewline
        \textit{DiffStk-MIRNN\textbf{(Ours)}} & $78.60$ & $99.50$ & $83.00$ & $97.50$  \tabularnewline
    \end{tabular}
    \caption{Percentage of correctly classified strings of various RNNs trained on the $D_3$ language (averaged over $10$ trials). We report the mean and best accuracy for each model. Test set used in all experiments had a mix of short samples and long ones of length up to$T = 102$ containing both positive and negative samples.}
    \label{D3performance}
\end{table}

\begin{table}[!t]
    \centering
    \begin{tabular}{l|r|r||r|r}
    \multicolumn{1}{l}{}&\multicolumn{2}{|c}{\begin{tabular}[x]{@{}c@{}}\textbf{$Train$}\\\end{tabular}}&\multicolumn{2}{|c}{\begin{tabular}[x]{@{}c@{}}\textbf{$Test$}\\\end{tabular}}\tabularnewline
    \multicolumn{1}{l}{\textbf{RNN Models}}&\multicolumn{1}{|c}{\begin{tabular}[x]{@{}c@{}}\textbf{Mean}\\\end{tabular}}&\multicolumn{1}{|c}{\begin{tabular}[x]{@{}c@{}}\textbf{Best}\\\end{tabular}}&\multicolumn{1}{|c}{\begin{tabular}[x]{@{}c@{}}\textbf{Mean}\\\end{tabular}}&\multicolumn{1}{|c}{\begin{tabular}[x]{@{}c@{}}\textbf{Best}\\\end{tabular}}\tabularnewline
\hline
        \textit{RNN} & $22.60$ & $25.00$ & $0.0$ & $0.0$  \tabularnewline
        \textit{LSTM} & $38.00$ & $42.80$ & $0.01$ & $0.04$  \tabularnewline
        \textit{GRU} & $25.00$ & $28.55$ & $0.01$ & $0.01$  \tabularnewline
        \textit{$\Delta$-RNN} & $32.62$ & $38.55$ & $0.02$ & $0.04$  \tabularnewline
        \textit{StackRNN} & $99.95$ & $100$ & $99.80$ & $100$  \tabularnewline
        \textit{DiffStk-RNN\textbf{(Ours)}} & $99.99$ & $100$ & $99.89$ & $100$  \tabularnewline
        \textit{DiffStk-LSTM\textbf{(Ours)}} & $99.80$ & $100$ & $98.00$ & $100$  \tabularnewline
        \textit{DiffStk-MRNN\textbf{(Ours)}} & $\textbf{100}$ & $\textbf{100}$ & $\textbf{100}$ & $\textbf{100}$  \tabularnewline
        \textit{DiffStk-MLSTM\textbf{(Ours)}} & $99.90$ & $100$ & $99.99$ & $100$  \tabularnewline
        \textit{DiffStk-MIRNN\textbf{(Ours)}} & $99.99$ & $98$ & $99.90$ & $100$  \tabularnewline
    \end{tabular}
    \caption{Percentage of correctly classified strings of various RNNs trained on the $D_6$ language (averaged over $10$ trials). We report the mean and best accuracy for each model. Test set used in all experiments had a mix of short samples and long ones of length up to$T = 102$ containing both positive and negative samples.}
    \label{D6performance}
\end{table}

\begin{table}[!t]
    \centering
    \begin{tabular}{l|r|r||r|r}
    \multicolumn{1}{l}{}&\multicolumn{2}{|c}{\begin{tabular}[x]{@{}c@{}}\textbf{$Train$}\\\end{tabular}}&\multicolumn{2}{|c}{\begin{tabular}[x]{@{}c@{}}\textbf{$Test$}\\\end{tabular}}\tabularnewline
    \multicolumn{1}{l}{\textbf{RNN Models}}&\multicolumn{1}{|c}{\begin{tabular}[x]{@{}c@{}}\textbf{Mean}\\\end{tabular}}&\multicolumn{1}{|c}{\begin{tabular}[x]{@{}c@{}}\textbf{Best}\\\end{tabular}}&\multicolumn{1}{|c}{\begin{tabular}[x]{@{}c@{}}\textbf{Mean}\\\end{tabular}}&\multicolumn{1}{|c}{\begin{tabular}[x]{@{}c@{}}\textbf{Best}\\\end{tabular}}\tabularnewline
\hline
        \textit{RNN} & $0$ & $0$ & $0$ & $0$  \tabularnewline
        \textit{LSTM} & $2.50$ & $5.45$ & $0$ & $0$  \tabularnewline
        \textit{GRU} & $0.9$ & $1.2$ & $0.3$ & $0.5$  \tabularnewline
        \textit{$\Delta$-RNN} & $0.5$ & $1.0$ & $0.01$ & $0.01$  \tabularnewline
        \textit{StackRNN} & $49.50$ & $100$ & $50$ & $100$  \tabularnewline
        \textit{DiffStk-RNN\textbf{(Ours)}} & $62.00$ & $100$ & $61.00$ & $100$  \tabularnewline
        \textit{DiffStk-LSTM\textbf{(Ours)}} & $63.50$ & $100$ & $62.20$ & $100$  \tabularnewline
        \textit{DiffStk-MRNN\textbf{(Ours)}} & $\textbf{64.00}$ & $\textbf{100}$ & $\textbf{63.00}$ & $\textbf{100}$  \tabularnewline
        \textit{DiffStk-MLSTM\textbf{(Ours)}} & $65.55$ & $100$ & $64.50$ & $100$  \tabularnewline
        \textit{DiffStk-MIRNN\textbf{(Ours)}} & $61.55$ & $99.99$ & $61.50$ & $99.99$  \tabularnewline
    \end{tabular}
    \caption{Percentage of correctly classified strings of various RNNs trained on the Palindrome language (averaged over $10$ trials). We report the mean and best accuracy for each model. Test set used in all experiments had a mix of short samples and long ones of length up to$T = 102$ containing both positive and negative samples.}
    \label{palindromeperformance}
\end{table}

\begin{table}[!t]
     \centering
    \begin{tabular}{l|r|r||r|r}
    \multicolumn{1}{l}{}&\multicolumn{2}{|c}{\begin{tabular}[x]{@{}c@{}}\textbf{$D_2$}\\\end{tabular}}&\multicolumn{2}{|c}{\begin{tabular}[x]{@{}c@{}}\textbf{$D_3$}\\\end{tabular}}\tabularnewline
    \multicolumn{1}{l}{\textbf{RNN Models}}&\multicolumn{1}{|c}{\begin{tabular}[x]{@{}c@{}}\textbf{n=120}\\\end{tabular}}&\multicolumn{1}{|c}{\begin{tabular}[x]{@{}c@{}}\textbf{n=160}\\\end{tabular}}&\multicolumn{1}{|c}{\begin{tabular}[x]{@{}c@{}}\textbf{n=120}\\\end{tabular}}&\multicolumn{1}{|c}{\begin{tabular}[x]{@{}c@{}}\textbf{n=160}\\\end{tabular}}\tabularnewline
\hline
        \textit{RNN} & $0.0$ & $0.0$ & $0.0$ & $0.0$  \tabularnewline
        \textit{LSTM} & $0.2$ & $0.0$ & $0.0$ & $0.0$  \tabularnewline
        \textit{GRU} & $0.2$ & $0.0$ & $0.0$ & $0.0$  \tabularnewline
        \textit{$\Delta$-RNN} & $0.0$ & $0.0$ & $0.0$ & $4.00$  \tabularnewline
        \textit{StackRNN} & $85$ & $50$ & $87$ & $60$  \tabularnewline
        \textit{DiffStk-RNN\textbf{(Ours)}} & $86.5$ & $53$ & $88.20$ & $60$  \tabularnewline
        \textit{DiffStk-LSTM\textbf{(Ours)}} & $83.20$ & $65$ & $85$ & $64.50$  \tabularnewline
        \textit{DiffStk-MRNN\textbf{(Ours)}} & $\textbf{92}$ & $\textbf{90}$ & $\textbf{91}$ & $\textbf{91}$  \tabularnewline
        \textit{DiffStk-MLSTM\textbf{(Ours)}} & $89.50$ & $85.00$ & $90$ & $89.99$  \tabularnewline
        \textit{DiffStk-MIRNN\textbf{(Ours)}} & $81.50$ & $50$ & $83.20$ & $58.50$  \tabularnewline
    \end{tabular}
    \caption{Percentage of correctly classified strings of various RNNs on $D_2$ and $D_3$ language when tested on longer strings for each model. The test set contained long samples of length up to $T = 160$. Note that we evaluated the best model on the longer strings}
    \label{longer1performance}
\end{table}

\begin{table}[!t]
    \centering
    \begin{tabular}{l|r|r||r|r}
    \multicolumn{1}{l}{}&\multicolumn{2}{|c}{\begin{tabular}[x]{@{}c@{}}\textbf{$D_6$}\\\end{tabular}}&\multicolumn{2}{|c}{\begin{tabular}[x]{@{}c@{}}\textbf{$Palindrome$}\\\end{tabular}}\tabularnewline
    \multicolumn{1}{l}{\textbf{RNN Models}}&\multicolumn{1}{|c}{\begin{tabular}[x]{@{}c@{}}\textbf{n=120}\\\end{tabular}}&\multicolumn{1}{|c}{\begin{tabular}[x]{@{}c@{}}\textbf{n=160}\\\end{tabular}}&\multicolumn{1}{|c}{\begin{tabular}[x]{@{}c@{}}\textbf{n=120}\\\end{tabular}}&\multicolumn{1}{|c}{\begin{tabular}[x]{@{}c@{}}\textbf{n=160}\\\end{tabular}}\tabularnewline
\hline
        \textit{RNN} & $0.0$ & $0.0$ & $0.0$ & $0.0$  \tabularnewline
        \textit{LSTM} & $0.0$ & $0.0$ & $0.0$ & $0.0$  \tabularnewline
        \textit{GRU} & $0.0$ & $0.0$ & $0.0$ & $0.0$  \tabularnewline
        \textit{$\Delta$-RNN} & $0.0$ & $0.0$ & $0.0$ & $0.0$  \tabularnewline
        \textit{StackRNN} & $92$ & $89$ & $65$ & $58$  \tabularnewline
        \textit{DiffStk-RNN\textbf{(Ours)}} & $94$ & $91$ & $69$ & $62.50$  \tabularnewline
        \textit{DiffStk-LSTM\textbf{(Ours)}} & $90$ & $89.20$ & $66$ & $64.50$  \tabularnewline
        \textit{DiffStk-MRNN\textbf{(Ours)}} & $\textbf{100}$ & $\textbf{99}$ & $\textbf{82.50}$ & $\textbf{80.00}$  \tabularnewline
        \textit{DiffStk-MLSTM\textbf{(Ours)}} & $98$ & $97.50$ & $77.50$ & $75.00$  \tabularnewline
        \textit{DiffStk-MIRNN\textbf{(Ours)}} & $91$ & $90.50$ & $67.50$ & $60$  \tabularnewline
    \end{tabular}
    \caption{Percentage of correctly classified strings of various RNNs on the $D_6$ and $Palindrome$ languages (averaged over $10$ trials) when tested on longer strings with mean and best accuracy for each model. The test set contained long samples up to length $T = 160$.}
    \label{longer2performance}
\end{table}

\subsection{Language Modeling}
\label{sec:lm}
In addition to CFGs, we evaluate all of our stack-augmented models on the Penn TreeBank language modeling task \cite{mikolovrnn10}. 
All models trained on this dataset consisted of $100$ hidden units and were trained over $50$ epochs using the Adam optimizer (also with patience scheduling for the learning rate). Dataset splits and settings was chosen based on prior related work \cite{joulin2015inferring}. Our models were able to match the performance of the LSTM and SRCN \cite{joulin2015inferring}.

\begin{table}[!t]
    \centering
    \begin{tabular}{l|r||r}
    \multicolumn{1}{l}{}&\multicolumn{1}{|c}{\begin{tabular}[x]{@{}c@{}}\textbf{$Validation$}\\\end{tabular}}&\multicolumn{1}{|c}{\begin{tabular}[x]{@{}c@{}}\textbf{$Test$}\\\end{tabular}}\tabularnewline
    \multicolumn{1}{l}{\textbf{RNN Models}}&\multicolumn{1}{|c}{\begin{tabular}[x]{@{}c@{}}\textbf{PPL}\\\end{tabular}}&\multicolumn{1}{|c}{\begin{tabular}[x]{@{}c@{}}\textbf{PPL}\\\end{tabular}}\tabularnewline
\hline
        \textit{RNN} & $137$  & $129$  \tabularnewline
        \textit{LSTM} & $120$ &  $\textbf{115}$  \tabularnewline
        \textit{SRCN} & $120$ & $\textbf{115}$  \tabularnewline
        \textit{StackRNN} & $124$ & $118$  \tabularnewline
        \textit{DiffStk-RNN\textbf{(Ours)}} & $122$ & $118$  \tabularnewline
        \textit{DiffStk-MRNN\textbf{(Ours)}} & $\textbf{119}$ & $\textbf{115}$  \tabularnewline
        \textit{DiffStk-MIRNN\textbf{(Ours)}} & $121$ & $117$  \tabularnewline
    \end{tabular}
    \caption{Perplexity of various model on Penn Tree bank. }
    \label{LMperformance}
\end{table}

\subsection{Operating on Longer Strings}
\label{sec:longer_strings}
If a model is operating in the same way as the equivalent correct pushdown automaton, in theory, it should be able to recognize a string of any length \cite{chomsky1959algebraic,chomsky1962context}. This provides an excellent way of analyzing the limitations of RNNs on longer strings. We created a separate test set containing $1500$ string samples of length ($105 < T$ and $T > 161$). This sampling of lengths is important in order to properly test and understand the limitations of RNNs, since interpretability is largely dependent on the fact that the RNN acts as a pushdown automaton (PDA). This is crucial if our aim is to extract a minimal PDA from the final, trained model weights \cite{das1993using,nndpa1998sun,weiss2017extracting,omlin1992secorderrnn,jacobsson2005rule}. 

\section{Results \& Discussion}
\label{sec:discussion} 
When testing on various CFGs, our experiments clearly show the importance of memory structure in RNNs. 
It is important to note that all stack-augmented RNNs were able to achieve full accuracy on the smaller test set (n$=$102) and a few models achieved $90$\%  accuracy on the larger test set(n $>$ 102). 
In Table \ref{D2performance}, all first order RNNs without memory appear to struggle to correctly recognize the $D_2$ grammar, where even the LSTM reaches only a 4\% accuracy. In contrast, the DiffStk-MRNN performed the best, which we hypothesize is due to its ability to approximate the higher order weights (operating similarly to tensor RNN). 
Similar performance was be observed for $D_3$, $D_6$, and the Palindrome grammar as demonstrated in Tables \ref{D3performance}, \ref{D6performance}, and \ref{palindromeperformance}. However, for the grammar $D_6$, as well as the palindrome, we observe a constant drop in performance as the string length increases, which indicates that even stack-augmented RNNs themselves have limitations. 

Based on our experiments and examination of the internal hidden state representations of the trained recognition models, we did find that whenever a test set contained more strings of longer length, the majority of RNNs struggle. This is largely due to the fact that the models just simply have not been exposed to strings of a long enough length during the training phase.
One way to combat this issue and improve model generalization would be to create a complex validation set and optimize the model based on its performance on that sample.Furthermore, one should use negative sampling when working with complex grammars, as our results demonstrate that RNNs that train with ``difficult'' negative samples generalize better to longer strings.
Finally, it is important to ensure that an RNN's weights do not change much when encountering a NO-OP (NoOP) operation. This avoids conflicts for whenever the weights of each stack operation are shifted by a small margin, since even a small degree of noise can alter the RNN's entire prediction.

\textbf{Continuous vs Discrete Stacks:} Using a continuous stack has been proven to very powerful when recognizing patterns generated by CFGs \cite{giles1992learning,joulin2015inferring,grefenstette2015learning}. However, it is still not entirely clear that continuous stack models are stable. Given that RNNs already in of themselves are difficult to train, adding a differentiable continuous stack introduces further instability during optimization, often resulting in sub-optimal performance. On the other hand, in theory, discrete stacks \cite{zeng1994discrete,mali2019neural} are much more interpretable and stable.
Future work should be directed towards combining discrete optimization with continuous optimization in order to solve tasks that might benefit from the use of a stable, discrete stack.

\section{Conclusion}
\label{sec:conclusion}
In this work, we introduced five improved stack-augmented recurrent neural network (RNN) models and evaluated their ability to recognize complex context free grammars (CFGs) with recurrence on long strings. 
We also developed various techniques to efficiently and stably train  differentiable stack-augmented models. To our knowledge, this is the first work to analyze the performance of a continuous stack on long strings for complex grammars. We showed that utilizing higher order weights/tensors do indeed improve model generalization. In addition, we demonstrated the value of memory structures and why they are important when learning from non-regular recursive language patterns. Since memory augmented RNNs suffer from stability issues, we provided an efficient, effective scheme to train them, especially for complex problems such as CFG recognition.
One interesting research direction would be to examine the effectiveness and stability of a discrete stack when the RNN that operates it is of first order. Furthermore, it is worth investigating if using efficient, viable alternatives to back-propagation through time, like that of \cite{ororbia2018continual}, might improve generalization and stability further. Another direction would be to explore data structures other than stacks and to develop the optimization algorithms needed to train networks to effectively operate these structures, especially on challenging real-world problems.

\bibliography{main_arxiv}
\bibliographystyle{acm}

\end{document}